\documentclass[runningheads]{llncs}
\usepackage{graphicx}
\usepackage{booktabs}
\usepackage{amsfonts}
\usepackage{multirow}
\usepackage{natbib}

\newcommand{\LL}{\mathcal{L}}
\newcommand{\GG}{\mathcal{G}}
\newcommand{\NN}{\mathbb{N}}
\newcommand{\RR}{\mathbb{R}}

\begin{document}
\title{ProcK: Machine Learning for Knowledge-Intensive Processes}
\author{Tobias Jacobs\inst{1}\orcidID{0000-0002-8130-4211} \and
Jingyi Yu\inst{1,2}\orcidID{0000-0002-1144-9765} \and \\ 
Julia Gastinger\inst{1}\orcidID{0000-0003-1914-6723} \and
Timo Sztyler\inst{1}\orcidID{0000-0001-8132-5920} 
}

\institute{
NEC Laboratories Europe GmbH, Heidelberg, Germany\\
\email{tobias.jacobs@neclab.eu}, \email{julia.gastinger@neclab.eu}, \email{timo.sztyler@neclab.eu} 
\and
Faculty of Electrical Engineering and Information Technology, RWTH Aachen\\
\email{jingyi.yu@rwth-aachen.de}
}
%
%\author{First Author\inst{1}\orcidID{0000-1111-2222-3333} \and
%Second Author\inst{2,3}\orcidID{1111-2222-3333-4444} \and
%Third Author\inst{3}\orcidID{2222--3333-4444-5555}}
%\institute{Princeton University, Princeton NJ 08544, USA \and
%Springer Heidelberg, Tiergartenstr. 17, 69121 Heidelberg, Germany
\maketitle              % typeset the header of the contribution
\begin{abstract}
We present a novel methodology to build powerful predictive process models. Our method, denoted \emph{ProcK (\textbf{Proc}ess \& \textbf{K}nowledge)}, relies not only on sequential input data in the form of event logs, but can learn to use a knowledge graph to incorporate information about the attribute values of the events and their mutual relationships. The idea is realized by mapping event attributes to nodes of a knowledge graph and training a sequence model alongside a graph neural network in an end-to-end fashion.This hybrid approach substantially enhances the flexibility and applicability of predictive process monitoring, as both the static and dynamic information residing in the databases of organizations can be directly taken as input data.
We demonstrate the potential of ProcK by applying it to a number of predictive process monitoring tasks, including tasks with knowledge graphs available as well as an existing process monitoring benchmark where no such graph is given. The experiments provide evidence that our methodology achieves state-of-the-art performance and improves predictive power when a knowledge graph is available.

\end{abstract}

%------------------

\section{Introduction}
\label{sec:intro}

We introduce \emph{ProcK (\textbf{Proc}ess \& \textbf{K}nowledge)}, a pipeline for predictive process monitoring. ProcK combines the usage of two complementary data representations in a novel way.

Predictive process monitoring deals with the task of forecasting properties of business processes that are currently under execution. This includes the type and occurrence time of future events as well as the process outcome. The primary input to predictive process models are logs recorded during process execution, and it is best practice in the process mining community to model them as sets of discrete \emph{events}. Each event is characterized by its case identifier, activity type, timestamp, and potentially further data; see \cite{van2011process}. 

Machine learning methods for predictive process monitoring described in literature can be separated into two main approaches. The more traditional approach is to hand-engineer 
a set of feature extraction functions that operate on top of event sequences.
The set of features, after some further pre-processing, is then used as input to train a machine learning model like e.g. an SVM (\cite{leontjeva2016complex}). The second approach relies on deep learning, feeding the raw event log directly into a deep neural network which builds meaningful features automatically during training. 
Because of the sequential nature of the input data, it is a natural choice to apply a recurrent neural network, like done by ~\cite{tax2017predictive}. More recently, feedforward networks have been demonstrated by ~\cite{mauro2019activity} and other authors to achieve superior performance in many cases.

Our work follows the deep learning paradigm, and, to the best of our knowledge, we are the first to complement the event sequence data model with an additional representation of the available input data as a knowledge graph. 
This idea is rooted in the first fundamental step of typical practical process mining projects: to extract the event log from the data lake of an organization (see~\cite{reinkemeyer2020process}), which is often structured in the form of one or more relational databases. When selecting data with the goal of building a high-quality prediction model, the limitations of the event sequence view become apparent: only a subset of the relevant data can be naturally expressed in the form of events with case identifier and timestamp.

\begin{example}
	\label{ex:1}
	For predicting the success of a loan repayment process at the time when the first rate has been paid, it might be relevant to take into account the bank account from which the rate was transferred. Specifically, the economic stability of the bank account's country might be an indicator.
\end{example}

In Example~\ref{ex:1}, the relevant piece of information ({economic stability}) is not an event, and it is neither a primary attribute of the event. It is rather an indirect attribute of the transfer event, which has to be derived via a specific semantic path (transfer $\to$ bank $\to$ country $\to$ economic stability). Domain experts could re-define such derived attributes by hand as primary event information, but this counteracts the benefit of deep learning to be applicable on top of the raw data. 

To address that issue, ProcK takes a knowledge graph as additional input. It stacks a sequence model for events on top of a graph neural network in order to compute meaningful event representations. In Example~\ref{ex:1}, information about the \emph{economic stability} can be propagated backwards across the path \emph{economic stability}$\leftarrow$ \emph{country} $\leftarrow$ \emph{bank} $\leftarrow$ \emph{transfer}, where \emph{economic stability, bank, country} are knowledge graph nodes and \emph{transfer} is a time-stamped event of a particular process. Having been integrated into the representation of \emph{transfer}, the information is then further processed by the event sequence model in order to predict the success probability of the repayment process.

In this work, we present the conceptual architecture of ProcK together with an implementation 
 based on deep learning models for graph-structured and sequential data. We also present results of an experimental study based on four datasets, three of them including a knowledge graph, from different application domains. Our experiments demonstrate that ProcK achieves state-of-the-art predictive performance, which is further improved by utilizing the additional knowledge graph input.

%To summarize our key contributions:
%\begin{itemize}
%\item We present a novel deep learning pipeline for predictive process mining, taking into account both sequential event data and graph-structured  knowledge.
%\item Our approach achieves state-of-the-art performance across a variety of domains
%We demonstrate the broad applicability and good predictive performance of our approach via an experimental study including several datasets.
 %\end{itemize}

\section{Related work}
\label{sec:relatedwork}

%Our approach combines ideas from several areas of related work, which we describe in the following subsections.

\subsection{Predictive process monitoring}

Our work presents a new approach for \emph{predictive process monitoring}, the task to predict future properties of processes from their execution logs. A considerable range of machine learning techniques have been studied in the context of process predictions. A review of seven methods that fall into the category of traditional machine learning (decision trees, random forests, support-vector machines, boosted regression, all with heavy feature engineering) has been published by~\cite{teinemaa2019outcome}.

Considering deep learning methods, due to the sequential nature of process logs, it is a straightforward approach to apply models designed for sequences.  \cite{tax2017predictive} study the usage of LSTM neural networks for various prediction tasks, including the next activities and the remaining process time. Further approaches based on RNN and LSTM networks have been presented by \cite{evermann2017predicting, tello2018predicting, camargo2019learning}. 

More recently, it has been demonstrated that feedforward networks often outperform recurrent neural networks for predictive process monitoring tasks. \cite{al2018predicting} employ 1D-convolutional networks, \cite{mauro2019activity} study stacked inception CNNs, and~\cite{pasquadibisceglie2019using} present a method where traces and their prefixes are first mapped onto a 2D image-like structure and then 2-dimensional CNNs are applied. Finally, \cite{taymouri2020predictive} present a prediction approach leveraging generative adversarial networks.

Although the main idea presented and evaluated in this work is independent from the particular choice of the sequence processing model, we share the experience of \cite{al2018predicting} and other authors that feedforward networks are more reliable to achieve good results in the process monitoring domain.

%Our approach is driven by the idea of constructing and using a knowledge graph as additional input source, which is independent from the choice of the particular network model for sequence processing. We have experimented with several network types, and we share the above experience that simpler feedforward models are more reliable to achieve good results in this domain.

A direction that has been followed by several researchers is to make use of explicit process models. More than two decades of research on process mining has yielded sophisticated algorithms to create graph-shaped models of processes from event log data, often in the form of Petri nets (\cite{van2011process}). From the viewpoint of such a process model, events trigger state changes of process instances, and predictive process monitoring models can take the current process state as input. \cite{van2011time} propose a solution to the problem of predicting the completion time, simply by calculating the mean remaining time for each state.  
Prediction from a combination of partial process models and event annotations has been performed by~\cite{ceci2014completion}. \cite{folino2014mining} combine some of the aforementioned ideas, first clustering events and processes to achieve more abstract process models and then applying cluster-specific prediction models.
Recently, \cite{theis2019decay} have presented a methodology to annotate states with time and other information, and use this information as input to a deep learning model. 

What makes the idea of using a graphical process model powerful is that it builds upon a mature topic in the process mining domain, where the graph can be constructed on top of the existing event log. Our work, in contrast, is built on the hypothesis that, in concrete applications, \emph{additional} non-sequential data is available, and that data is naturally modeled as a graph. In other words, process modeling is a specific type of feature pre-processing, while our approach taps a previously unused resource of data.

\subsection{Combined sequence and graph models}

%Our work combines input from sequences with graph-structured input into a single model. 
The machine learning model we employ in our work uses a combination of graph-structured and sequential data as input.
To our knowledge, we are the first to apply this approach to the domain of predictive process monitoring. Nevertheless, the benefits of combining such complementary views have been demonstrated in other domains. \cite{fang2018mobile} construct a graph based on spatial distances between mobile cells. Their model for cellular demand prediction uses an LSTM to compute a feature vector for each cell based on its demand history, then a graph convolutional network is used to model influences between nearby cells. \cite{hu2019graph} address the problem of predicting the freezing-of-gaits symptom of Parkinson disease patients from video segments, where the first layer of their model is used to compute a representation of anatomic joints and their interactions as a graph. The authors introduce specialized LSTM cells to model both time-based interactions between subsequent video segments and interactions between joints. \cite{wang2020advanced} combine the graph-based representation of molecules with the SMILES string representation. Two individual models are trained for those two input representations, and the models are combined using ensemble techniques. An application in the retail domain has been proposed by~\cite{chang2021sequential}. Here the first step is to convert the sequence of past user interactions with items into a graph with nodes representing items, and graph-based models are then applied to predict the interest of users.

While a variety of approaches to combine graphical and sequential input have been proposed in the body of work mentioned above, our approach has some unique and novel features. 
Firstly, each of the above methods performs some sort of node or graph classification which is only assisted by the sequential input. In our work, event sequences constitute the primary input, and the knowledge graph is used to assist the model to interpret the event data.
Secondly and most importantly, the knowledge graph is used in our work to capture overall knowledge about the domain and application context, while specific information about each instance of the prediction problem is represented by its event sequence.

%we use a  single knowledge graph that is shared throughout all instances of the prediction problem at hand, while each particular instance is characterized by its own event sequence. Thus, the graph represents generic knowledge about the domain and application context. 

%\subsubsection{Temporal knowledge graphs}

Another direction of related work is machine learning for dynamic knowledge graphs, called temporal knowledge graphs, where nodes, attributes, or relations change over time. The area of temporal knowledge graph reasoning can be divided into the \emph{interpolation} and the \emph{extrapolation} setting. As described by \cite{jin2020recurrent}, in the interpolation setting, new facts are predicted for time steps up until the current time step, taking into account time information from past and current time steps; see \cite{garcia2018learning}. The methods in the extrapolation setting predict facts for future time steps. %, for example with the goal of (future) link prediction. 
Recent work in the extrapolation setting includes %xERte [TODO], 
RE-NET by \cite{jin2020recurrent} and CluSTeR by \cite{li2021search}. %, who both represent each time step with a so-called graph snapshot including all triples that are true at this time step

%CluSTeR aims to predict future facts with a two-stage approach: clue searching and temporal reasoning.
%% (1) searching clue paths related to a given query using reinforcement learning to learn a beam search policy, and (2) temporal reasoning using a graph convolution network based sequence method. 
%RE-NET employs a recurrent event encoder to encode past facts, and a neighborhood aggregator to model the connection of facts at the same timestamp to infer, based on those two modules, future facts in a sequential manner.
%RE-NET and CluSTeR both use the notion of events: they define an event as a timestamped edge, i.e., (subject entity, relation, object entity, time) and describe every fact that occurs at a certain timestamp as an event.
%% xERte, presented by [TODO] reasons over query-relevant subgraphs of the temporal knowledge graph
%% Here the target of prediction models are future properties of graph elements. TODO include 2-3 references and their descriptions. Some %articles on this topic use a notion of events to represent changes in the knowledge graph that take place at a specific time. 

The predictive process management application in our work requires to treat events as independent input. Modeling each event as a knowledge graph element would be technically possible, but only at the price of scalability, as typical applications include tens or hundreds of millions of events. Thus, in our work events do not have a direct interpretation as nodes or edges, but they instead contain attributes in the form of references to graph nodes. Conversely, triples in our knowledge graphs are in general not necessarily interpretable as events.

To summarize the discussion of our work in light of state-of-the-art, we are the first to utilize a knowledge graph as additional input to predictive process monitoring models to help interpreting the event data. The idea is realized by a new type of neural network architecture which takes events as primary input and learns to utilize an additional knowledge graph to interpret the event data.

\section{Preliminaries}
\label{sec:datamodel}

Following the process mining terminology, 
an \emph{event log}  $\LL = (L,C,T,A)$ consists of an event set $L$, a set $C$ of cases, a set $T$ of possible event types, and a set~$A$ of additional event attributes. Each event $\ell = (c_\ell,t_\ell,\tau_\ell,\alpha_\ell) \in L$ is a 4-tuple characterized by its case identifier $c_\ell \in C$, its event type $t_\ell \in T$, a timestamp $\tau_\ell \in \NN$, and a partial assignment function $\alpha_\ell:A \to V$ which specifies the values of a subset of attributes. 
%In practice, the set of attributes of an event will often be determined by the event type. 
For each case $c \in C$ we define $L_c := \{\ell \in L \mid c_\ell = c \}$ as the subset of events belonging to case $c$.

The domain $V$ of possible attribute values is arbitrary in general; in this work it will be assumed that $V$ is the node set of a knowledge graph. This assumption represents only a mild limitation, because categorical attribute values that do not appear in the given knowledge graph can simply be interpreted as isolated nodes. In fact, our experimental study includes one dataset where no knowledge graph is given at all. Extending ProcK with the ability to incorporate numerical attributes \emph{directly} (i.e. without discretizing them to categorical attributes) remains for future work.

A knowledge graph (also called \emph{knowledge base}) $\GG = (V,R,E)$ is a directed graph defined by the node set $V$, relation types $R$, and edges $E$, where each edge $(v,r,v^\prime) \in E$ is a triple containing the head node $v \in V$, relation type $r \in R$, and tail node $v^\prime \in V$.

As mentioned above, in typical practical applications, the input data will originate from the databases of the organization that performs predictive process monitoring. Relational databases are likely to contain time-stamped records that can become events. At the same time, mutual references between different tables are a fundamental element of relational data models, which makes it straightforward to interpret a part of the database as a knowledge graph.
% where table rows become inter-linked nodes. 
For the purpose of our experimental study we have developed tools to extract both event data and a knowledge graph from a database dump.

\section{ProcK architecture}
\label{sec:architecture}

We first specify, in Section~\ref{sec:highlevel}, the conceptual architecture which is composed of four functional components $\textrm{GNN}, f, F,$ and $\textrm{SM}$. Then, in Section~\ref{sec:implementation}, our implementation of each of the components is described.

\subsection{Conceptual architecture}
\label{sec:highlevel}

\begin{figure}[t]
\begin{center}
\includegraphics[width=0.8\linewidth]{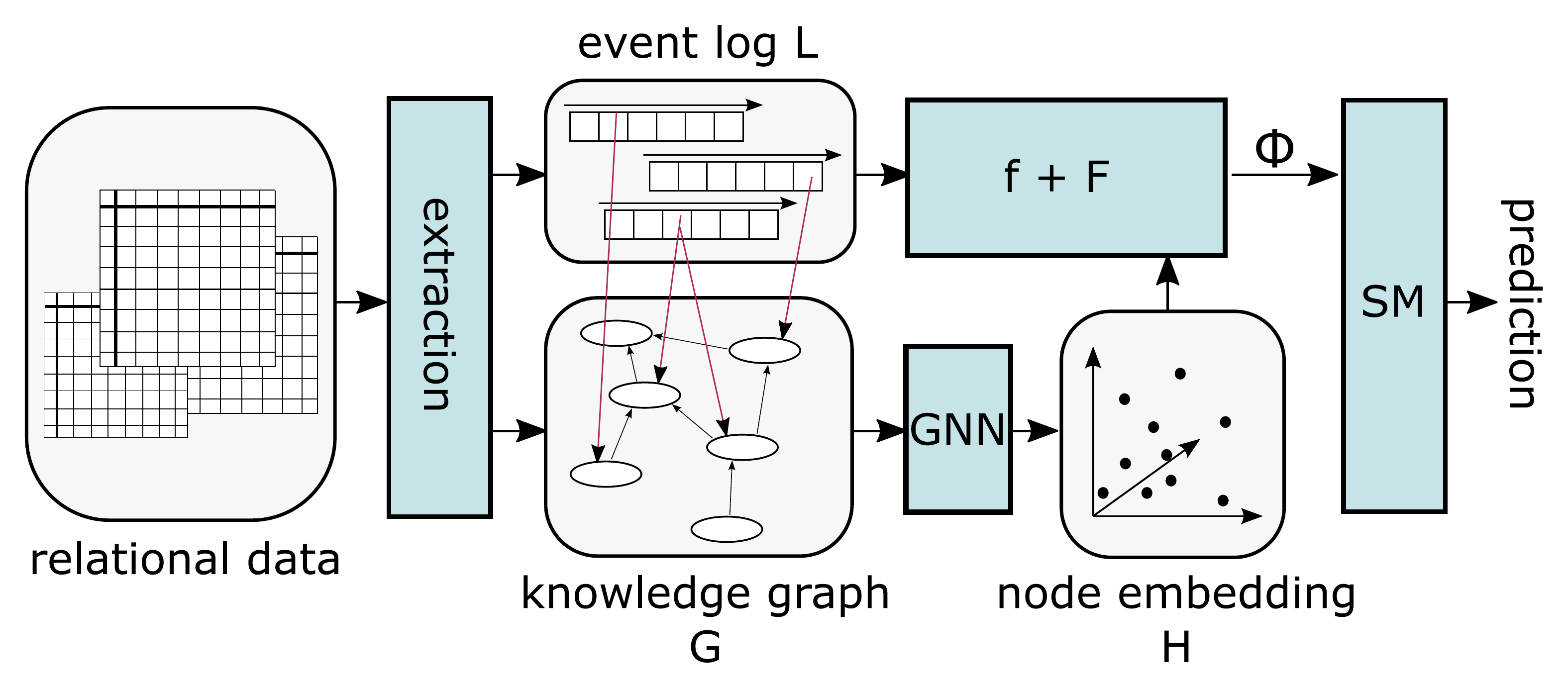}
\end{center}
\caption{
ProcK conceptual architecture. From relational or tabular data, an event log and a knowledge graph is extracted. Then, the four components $\textrm{GNN}, F, f,$ and $\textrm{SM}$ of the neural network model subsequently compute node embedding $H$, event embeddings $\Phi$, and the final prediction.
\label{fig:pipeline}
}
\end{figure}

The architecture of ProcK, depicted in Figure~\ref{fig:pipeline}, combines a graph neural network with a model for sequential data. From the bottom to the top of the network, input elements like nodes, edges, timestamps, and events will be encoded by embedding vectors. We employ a fixed embedding dimensionality $d \in \NN$ across all layers. %For the sake of readabilityIn the descriptions below we omit the parameterization vectors of the individual functions for better readability of the formulae.

The first component of the ProcK model is a graph neural network $\textrm{GNN}$ which computes embedding vectors $H = (h_v)_{v \in V}$ containing an embedding $h_v \in \RR^d $ for every node $v$ of the knowledge graph. Formally,
\begin{equation}
	h_v := \textrm{GNN} (\GG,v) \ , v \in V \ ,
\end{equation}
where  $\GG = (V,R,E)$ is the given knowledge graph.

Next, consider a single event $\ell = (c_\ell,t_\ell,\tau_\ell,\alpha_\ell)$ from the event log $\mathcal{L}$. Let $F$ be an aggregation function for sets of $d$-dimensional vectors. The first step of constructing the event embedding is to compute
\begin{equation}
	\beta_\ell := F(\{h_v \mid v \in \alpha_\ell(A)\}) \in \RR^d \ .
\end{equation}
Recall from the previous section that $\alpha_\ell$ is a partial assignment function which specifies values for a subset of the attributes from A. In the above equation, $a_\ell(A)$ is the set of those attribute values.

%where $\alpha_\ell(A)$ is the set of $\ell$'s attribute values. 

Further, we employ a timestamp embedding function $f:\NN \to \RR^d$ and compute the final event embedding as 
\begin{equation}
	\phi_\ell := \beta_\ell + f(\tau_\ell) \ .
\end{equation}

Having computed the sequence representation as a series of $d$-dimensional vectors $\Phi = (\phi_\ell)_{\ell \in L_c}$ for the given case $c \in C$, we feed the event representations into a sequence model $\textrm{SM}$ to compute the prediction for the given case:
\begin{equation}
	P_c := \textrm{SM}(\{\phi_\ell \mid \ell \in L_c\}) \ .
\end{equation}   

The structure of $P_c$ depends on the prediction target; e.g. it can be a single real number for regression tasks or a vector of probabilities for a classification task. All four functions $\textrm{GNN},F,f,\textrm{SM}$ are potentially parameterized by trainable vectors $\Theta_\textrm{GNN}, \Theta_F, \Theta_f,\Theta_\textrm{SM}$, respectively.

\subsection{Implementation}
\label{sec:implementation}

The bottom layer of our $\textrm{GNN}$ implementation is a trainable embedding vector $h^0_v \in \RR^d$ for each node $v \in V$, as well as an embedding vector $h^0_r \in \RR^d$ for every relation $r \in R$. We compute higher-level node embeddings using graph convolution layers, where we adopt a simplified version of the compGCN architecture proposed by \cite{vashishth2019composition} as described below. Given the layer~$i$ embeddings $(h^i_v)_{v \in V}, (h^i_r)_{r \in R}$, two transformations are applied for each $v \in V$:
\begin{equation}
	h_{v,\textrm{self}}^{i+1} := W^{i+1}_\textrm{self} \cdot h^i_v
\end{equation}   
\begin{equation}
\label{eq:backward}
 h_{v,\textrm{adj}}^{i+1} := W^{i+1}_\textrm{adj} \cdot \sum_{(v,r,v^\prime) \in E} \textrm{cmp}(h^i_r,h^i_{v^\prime}) \ ,
\end{equation}
where $\textrm{cmp}:\RR^d \times \RR^d \to \RR^d$ is the \emph{composition operator}. Our implementation supports the composition operators of addition and element-wise multiplication; we employ the latter throughout our experiments. The node and relation embeddings on layer $i+1$ are then computed via
\begin{equation}
	h^{i+1}_v = \textrm{relu}( h_{v,\textrm{self}}^{i+1} + h_{v,\textrm{adj}}^{i+1}) \ \ , v \in V
\end{equation}  
\begin{equation}
	h^{i+1}_r = W^{i+1}_\textrm{rel} \cdot h^i_r \ \ , r \in R \ .
\end{equation}
The computations on each layer are parameterized by $W^{i+1}_\textrm{self}, W^{i+1}_\textrm{adj}, W^{i+1}_\textrm{rel} \in \RR^{d \times d}$.  Equation~\ref{eq:backward} specifies backward flow across the edges of the directed graph. In \cite{vashishth2019composition}, forward flow with independent parameterization is additionally specified. 
%bi-directional information flow along the directed edges is also specified, and the forward flow is parameterized independently of the backward flow. Forward flow is 
This is also supported by our implementation, but we only consider backward flow in our experiments. The reason is that the knowledge graphs in our datasets contain nodes with a huge number of incoming links, and we experienced that summing up over them during the calculation of forward flow de-stabilizes the model and does not scale well. 

Having computed $k$ layers of graph convolution, the final node embeddings are the GNN output:
\begin{equation}
	h_v = \textrm{GNN} (\GG,v) := h_v^k \ \ , v \in V \ .
\end{equation}
For the aggregation function $F$ we employ mean pooling across the nodes referenced in each event:
\begin{equation}
	\beta_\ell = F(\{ h_v \mid v \in \alpha_\ell(A) \}) := \frac{1}{|\alpha_\ell(A)|} \sum_{v \in \alpha_\ell(A)} h_v \ .
\end{equation}
For the timestamp embedding function $f$ our implementation supports parameterized embedding and non-parameterized embedding based on sinusoids as described by \cite{vaswani2017attention}. We have however found that, for the prediction tasks included by our experimental study, the timestamp input is not essential; thus we applied the constant zero function there for most datasets.

We now describe our implementation of the sequence model $\textrm{SM}$.  First, a linear transformation is applied to the embeddings $(\phi_\ell)_{\ell \in L_c}$ :
\begin{equation}
	\phi^\prime_\ell = W_1 \cdot \phi_\ell \ \ , \ell \in L_c \ .
\end{equation}
After this initial transformation, we aggregate over the events of the sequence using mean pooling:
\begin{equation}
	\phi^{\prime\prime} = \frac{1}{|L_c|} \sum_{\ell \in L_c} \phi^\prime_\ell \ .
\end{equation}
A final fully connected hidden layer connects the aggregated events with the output:
\begin{equation}
	\phi^{\prime\prime\prime} = \textrm{relu}\left(W_2 \cdot \phi^{\prime\prime}\right) \ ,
\end{equation}
\begin{equation}
	P_c = \textrm{SM}(\{\phi_\ell \mid \ell \in L_c\}) := g(W_3 \cdot \phi^{\prime\prime\prime}) \ .
\end{equation}
The dimensionality of the matrices is $W_1,W_2 \in \RR^{d \times d}$ and $W_3 \in \RR^{o \times d}$. For binary classification problems, $o=1$, and $g$ is the sigmoid activation function. For multi-class classification, $o$ corresponds to the number of classes and $g$ is the softmax function. Finally, for regression problems, $o=1$ and $g$ is the identity function.

We remark that the design choice of the sequence model is a result of exploratory experiments with various architectures. During those experiments we observed that, throughout the datasets, more sophisticated architectures (recurrent networks, transformer) did not lead to better results and introduced stability problems. This is in line with the finding, reported by \cite{al2018predicting} and \cite{mauro2019activity}, that feedforward networks outperform the LSTM architecture for predictive process monitoring tasks.

\section{Experiments}
\label{sec:experiments}

\subsection{Data}

Our experiments encompass six prediction tasks using four datasets; see Table~\ref{tab:datasets} for a summary.

\begin{table}[t]
\begin{center}
\begin{tabular}{|l|c|c|c|}
\hline
{dataset}  & {knowledge graph} & {event log} & {reference} \\
\hline
{OULAD} & {240K nodes, 1.1M edges} & {33K cases, 11M events} & {\cite{kuzilek2017open}}\\
{PKDD99}  & {200K nodes, 1,1M edges} & {4.5K cases, 1M events} & {\cite{Financial}}\\
{BPI12}  & { -- } & {13K cases, 180K events} & {\cite{vandongen_2012}}\\
{DBLP}  & {370K nodes, 1.2M edges} & {29K cases, 1.7M events} & {\cite{tang2008arnetminer}}\\
\hline
\end{tabular}
\end{center}
\caption{
\label{tab:datasets}
Summary of the event logs and knowledge graphs extracted from the datasets.
}
\end{table}

\begin{figure}
\includegraphics[width=\linewidth]{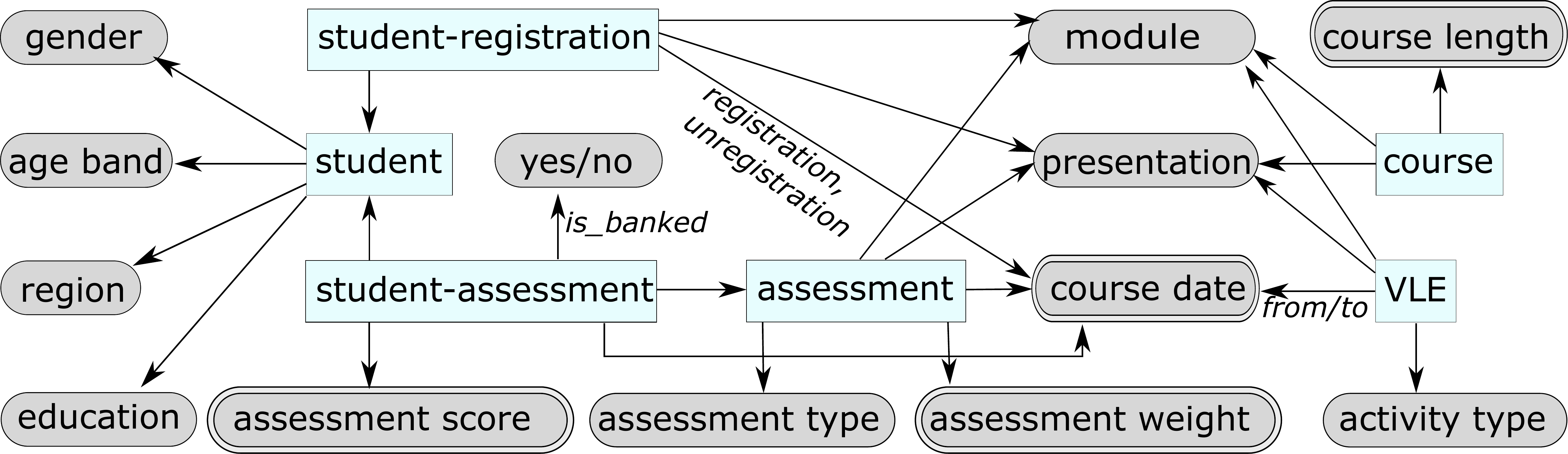}
\caption{
Schema of the knowledge graph generated from the OULAD dataset.  Box-shaped meta-nodes represent nodes generated from table rows, single-lined oval meta-nodes originate from categorical values, and double-lined meta-nodes represent discretized numerical values. Each arrow represents a distinct relation type (sometimes two); annotations have been added only where the type is not self-explanatory.
\label{fig:oulad_kb_structure}
}
\end{figure}

Three tasks are based on the Open University Learning Analytics (OULAD) dataset provided by~\cite{kuzilek2017open}. The dataset has the structure of a relational database dump, consisting of seven tables that represent information about students registering for courses, interacting with the study material, taking assessments and exams. From this data we extracted a knowledge graph as described in Figure~\ref{fig:oulad_kb_structure}. 
We further extracted an event log where each case $c \in C$ represents one student taking part in one course. There are five event types: \emph{case info}, containing links to student, module, and semester, %presentation (in this dataset, the presentation is the semester where the course module takes place)., 
%As such events represent information about the case as a whole, they are always assigned timestamp 0 and appear as the first event of the sequence.
	\emph{assessment}, containing the assessment submission date as the timestamp, and a link to the corresponding student assessment node in the knowledge graph, 
	\emph{student registration} and \emph{deregistration}, containing the date of (de)registration as the timestamp, and a link to the student registration node in the knowledge graph,
and finally
	\emph{VLE interaction}, containing as timestamp the time of interaction with material of the Virtual Learning Environment (VLE), as well as a link to the knowledge graph node representing the material.

For the OULAD dataset we distinguish between three prediction targets: \emph{dropout} (predict whether the student will drop out from the course), \emph{success} (predict whether the student will finish the course successfully), and \emph{exam score} (regression task to predict the final exam score, a number between 0 and 100). We further consider three different time horizons: \emph{late prediction} is a variant of the prediction task where all events that happened during the course (except for the final exam and the deregistration event) are available as input. 
%We have however decided to generally not use the VLE interaction events; they outnumber the other events by orders of magnitude and we found that the generalizability of the models suffer when using them as input. 
\emph{Early prediction} refers to predictions taking into account only the events that happened before the 60th day of the course (all course modules take between 234 and 269 days), while in \emph{very early prediction} no events have been recorded and only the  \emph{case info} is available. Across all tasks and their variants, we uniformly selected 20\% of the data for validation and 5\% for testing, like done by \cite{jha2019oulad}.
%Timestamps in the raw dataset are given in days since the start of the course, and we keep this time resolution in the model input. However, as some timestamps in the original dataset are negative (students work on a course before it officially starts), we add a positive constant to all timestamps such that they become non-negative.

%We randomly selected 5\% of the cases as the test set. The remaining 

A dataset from the financial domain was provided by~\cite{Financial} for the \emph{PKDD99} challenge. This dataset also comes in the form of multiple inter-connected tables, representing bank accounts, financial transactions, clients, geographical districts, and loans. The task is to predict the status of a loan (noncritical or critical) given the history of transactions and information about the loan and the client. The triples of the knowledge graph relate loans to accounts, accounts to districts, and bank orders as well as transactions to banks. Categorical and numerical attributes (after discretization) of those entities are additionally represented as neighbor nodes of them. The two event types extracted from the dataset are \emph{case info}, including links to the loan and the account node, and \emph{transaction}, containing a link to the node representing the transaction in the knowledge graph. We chose 20\% of the data for validation and another 20\% for testing. Because the number of samples in this dataset is comparably small and the dataset is rather imbalanced with only 10\% of the loans having a critical status, we used stratified sampling to enforce the same balance of positive and negative examples in the training, validation, and test set.

Our second dataset from the financial domain represents an established process mining benchmark, but it comes only with an event log. The events contained in the \emph{BPI12} dataset (\cite{vandongen_2012}) were recorded during the application procedure at a financial institution. Due to the lack of a knowledge graph, we treat all event attributes as isolated nodes of a graph without edges. To make our experiments comparable to~\cite{tax2017predictive}, we only consider events that are marking the completion of manually executed subprocesses, and we used the latest 30\% among the sequences as the test set. Another 20\% of the sequences were chosen as the validation set uniformly at random. As an additional pre-processing step, we treat every prefix of every sequence as one sample where the task is to predict the type of the next event. This step was done separately for the training, validation, and test set.

Our final prediction problem is the number of future citations of papers published in the year 2000 as reported in the DBLP dataset introduced by~\cite{tang2008arnetminer}. We extracted a knowledge graph consisting of relations between papers, authors, and venues. To prevent information leakage, events related to a publication after 2000 are not considered for the knowledge graph construction. For each published paper, the history of previous publications of all authors is used as the sequence of events. We uniformly selected 20\% of the data for validation and 5\% for testing.

\subsection{Setup}

The machine learning models were implemented in Tensorflow 2.5.0, and the computational experiments were performed on GPUs (Nvidia GeForce GTX 1080 Ti). For all classification tasks we used the cross-entropy loss function for training and selected the model having the highest accuracy on the validation set among all 200 training epochs. The learning rate was set to 0.01, and we chose an embedding width of $d=100$. We also applied dropout and l2-regularization, which required different strategies for different tasks. Our implementation supports dropout with uniform rate after each aggregation layer in the graph convolutional network and after the final fully-connected dense layer. Table~\ref{tab:reg} lists the chosen strategy used for ProcK for every task, as well as for the LSTM baseline, where we found a different configuration to work best. 

\begin{table}[tbp]
\begin{center}
\begin{tabular}{|c|c|c|c|c|}
\hline
{task} & {dropout rate} & {l2-weight} & {GC layers} & {time embedding}\\
\hline
{OULAD (dropout)} & {0.1} & {0.01} & {1} & {none}\\
{OULAD (success)} & {0.1} & {0.01} & {1} & {none} \\
{OULAD (score)} & {0.7} & {0} & {1}& {none}\\
{PKDD99} & {0.1} & {0.03} & {3}& {none}\\
{BPI12} & {0.1} & {0.01} & {1}& {parameterized}\\
{DBLP} & {0.5} & {0} & {0}& {none}\\
\hline
{all tasks, LSTM model} & {0.25} & {0.01} & {-} & {implicit}\\
\hline
\end{tabular}
\end{center}
\caption{
\label{tab:reg}
Hyperparameters used in the experiments. Note that the LSTM model implicitly takes into account the event position by design.
}
\end{table}

%\begin{table}
%\begin{center}
%\begin{tabular}{rccc}
%\toprule
%{} & {dropout} & {success} & {score} \\
%\midrule
%transformer blocks	& {3} & {3} & {2}\\
%{transformer heads}	& {2} & {2} & {2} \\
%{event embedding} & {plain} & {TB} & {TB} \\
%{TB hidden units}	& {100} & {100} & {100} \\
%{pred. layer hidden units} & {100} & {100} & {100} \\
%{embedding dimension} & {100} & {100} & {100} \\
%{dropout rate}	& {0.1} & {0.1} & {0.1} \\
%{optimizer}	& {Adam} & {Adam} & {Adam} \\
%{learning rate} & {0.0005} & {0.0005} & {0.001} \\
%{training batch size}	& {64} & {64} & {64} \\
%{training epochs}	& {200} & {200} & {200} \\
%\bottomrule
%\end{tabular}
%\end{center}
%\caption{
%\label{tab:hyperparameters}
%Hyperparameters of the event sequence model for the three prediction targets.
%}
%\end{table}

\subsection{Results}

\begin{table}
\begin{center}
\begin{tabular}{|c|c|c|c|}
\hline
{\textbf{prediction task}} & {\textbf{model}}  & {\textbf{accuracy}} & {\textbf{AUC}} \\
\hline
{OULAD (dropout, late)} & {ProcK} & { 0.86} & {\bf 0.93}\\
{} & {ProcK (no KG)} & { 0.86} & {\bf 0.93}\\
{} & {LSTM} & { 0.86} & {0.92}\\
{} & {GBM (\cite{jha2019oulad})} & {-} & {0.91}\\
\hline
{OULAD (dropout, early)} & {ProcK} & \bf {0.83} & {\bf 0.84}\\
{} & {ProcK (no KG)}& {0.81} & {0.82}\\
{} & {LSTM}  & {0.81} & {0.82}\\
\hline
{OULAD (dropout, very early)} & {ProcK} & {0.68} & {0.58}\\
{} & {ProcK (no KG)}  & {0.69} & {\bf 0.60}\\
{} & {LSTM}& {\bf 0.70} & {0.57}\\
\hline
{OULAD (success, late)} & {ProcK} & {0.87} & {0.91}\\
{} & {ProcK (no KG)} & {\bf 0.88} & {0.88}\\
{} & {LSTM}  & {0.86} & {0.86}\\
{} & {GBM(\cite{jha2019oulad})} & {-} & {\bf 0.93}\\
\hline
{OULAD (success, early)} & {ProcK} & {0.73} & {\bf 0.74}\\
{} & {ProcK (no KG)}  & {0.73} & {0.73}\\
{} & {LSTM} & {\bf 0.75} & {0.73}\\
\hline
{OULAD (success, very early)}  & {ProcK} & {\bf 0.69} & {\bf 0.58}\\
{} & {ProcK (no KG)} & {\bf 0.69} & {0.56}\\
{} & {LSTM} & {0.68} & {0.57}\\
\hline
{PKDD99} & {ProcK}  & {\bf 0.89} & {\bf 0.71}\\
{} & {ProcK (no KG)}  & {\bf 0.89} & {\bf 0.71}\\
{} & {LSTM} & {\bf 0.89} & {0.50}\\
\hline
{BPI12 (KG not available)} & {ProcK} & {\bf 0.83} & {-}\\
{} & {LSTM}& {0.71} & {-}\\
{} & {LSTM(\cite{tax2017predictive})}& {0.76} & {-}\\
\hline
\hline
{\textbf{prediction target}} & {\textbf{model}}  & \multicolumn{2}{|c|}{RMSE}\\ % {\textbf{RMSE}} & {} \\
\hline
{OULAD (score, late)} & {ProcK} & \multicolumn{2}{|c|}{\bf 18.93}\\
{} & {ProcK (no KG)} & \multicolumn{2}{|c|}{18.95}\\
{} & {LSTM} & \multicolumn{2}{|c|}{20.35} \\
\hline
{OULAD (score, early)} & {ProcK} & \multicolumn{2}{|c|}{\bf 19.88}\\
{} & {ProcK (no KG)}  & \multicolumn{2}{|c|}{\bf 19.88}\\
{} & {LSTM}  & \multicolumn{2}{|c|}{21.08}\\
\hline
{OULAD (score, very early)} & {ProcK}  & \multicolumn{2}{|c|}{\bf 20.10}\\
{} & {ProcK (no KG)}  & \multicolumn{2}{|c|}{20.44}\\
{} & {LSTM} & \multicolumn{2}{|c|}{20.13}\\
\hline
{DBLP}  & {ProcK}  & \multicolumn{2}{|c|}{\bf 3.98}\\\
{}  & {LSTM} & \multicolumn{2}{|c|}{4.01} \\
\hline
\end{tabular}
\end{center}
\caption{
\label{tab:results}
Experimental results
}
\end{table}

The results of our experiments are displayed in Table~\ref{tab:results}. The top part of the table contains results for classification problems, where the accuracy and the Area Under the Curve (AUC, only for binary classification problems) metric are reported. On most datasets a knowledge graph is available, and we compare models trained with it to models trained without. As a baseline we also evaluated an LSTM model which was trained using the event sequence as input. Whenever available, the table also contains the best results reported in literature.

For the AUC metric, it turns out that the availability of additional data in form of a graph improves the ability of the model to correctly separate positive and negative test samples, both in comparison with ProcK without knowledge graph and the LSTM model. The improvement is observable across most of the problems. For dropout prediction on the OULAD dataset, all three deep learning models outperform the results of Gradient Boosting Machine (GMB), reported by \cite{jha2019oulad}, while GBM performs better for the success prediction task. 

When looking at the accuracy metric, the advantage of the knowledge graph input is not that clearly visible; only on two out of seven problems the full ProcK model with knowledge graph exhibits the best performance,  on one problem it is outperformed by ProcK without knowledge graph input, and two tasks the LSTM model performs best. The only classification problem with more than two classes is next event type prediction on the BPI12 dataset. Here no knowledge graph is given, and ProcK outperforms the LSTM model by a large margin. This finding is consistent to results found in other works e.g. by \cite{mauro2019activity}.

We are the first to study variants of OULAD with different points of prediction time (early and very early prediction). It turns out that the length of the event log makes a significant difference, with AUC values decreasing to less than 0.60 when only the initial case information is available. However, the benefit of using knowledge graph input does not seem to depend on the length of the event log, which can be explained with the fact that less event input also means less information from the knowledge graph.

The bottom part of Table~\ref{tab:results} contains the results for regression problems, including exam score  (OULAD dataset) and number of citations (DBLP dataset). We do not include ProcK without knowledge graph input for the latter, because here the number of graph convolution layers is set to zero (see Table~\ref{tab:reg}), making models with and without knowledge graph input equivalent. The table reports the root mean square error, and here again the benefits of the knowledge graph input  can be demonstrated across the variants of the score prediction task.

\section{Summary and conclusion}
\label{sec:conclusion}

In this work, we introduced ProcK, a novel machine learning pipeline for data from knowledge-intensive processes. Within the pipeline, two complementary views of the available information are first extracted from raw tabular data  and then re-combined as input to the downstream prediction model. We implemented prototypes of each pipeline component, and we tested their interplay on six prediction tasks on four datasets. We could demonstrate on the majority of classification tasks that ProcK achieves improved AUC values when having a knowledge graph available as input, but further investigation of the accuracy metric remains a task for future work. Also for regression tasks ProcK exhibits a small but consistent advantage in terms of the RMSE metric.

There are several additional directions for future work. One interesting question from a practical viewpoint is how machine learning can be employed to extract the knowledge base and event log from the source databases in an automatic and configuration-free manner. Furthermore, while ProcK has been applied for prediction tasks so far, the ability to make process recommendations (e.g. events that should happen in the future to positively influence the outcome of cases) will be an important next step. 

\paragraph{Ethics discussion} All experiments reported in this work are based on anonymized datasets (OULAD, PKDD, BPI12) or data actively published by the data subjects (DBLP). Nevertheless, the presented technology is applicable to ethically sensitive tasks, including assessment of loan applications and performance prediction of humans. A careful assessment of potential ethical issues has to be carried out prior to bringing this work to application. 

%Authors may include a statement of the potential broader impact of their work, including its ethical aspects and future societal consequences. This part can be put in either the main body of the paper or on the reference page. It is optional but is highly recommended for papers working with sensitive data or on sensitive tasks.

\bibliographystyle{named}
\bibliography{ProcK}

\begin{thebibliography}{}

\bibitem[\protect\citeauthoryear{Al-Jebrni \bgroup \em et al.\egroup
  }{2018}]{al2018predicting}
Abdulrhman Al-Jebrni, Hongming Cai, and Lihong Jiang.
\newblock Predicting the next process event using convolutional neural
  networks.
\newblock In {\em 2018 IEEE International Conference on Progress in Informatics
  and Computing (PIC)}, pages 332--338. IEEE, 2018.

\bibitem[\protect\citeauthoryear{Berka}{1999}]{Financial}
Petr Berka.
\newblock {Workshop notes on Discovery Challenge PKDD'99}.
\newblock 1999.

\bibitem[\protect\citeauthoryear{Camargo \bgroup \em et al.\egroup
  }{2019}]{camargo2019learning}
Manuel Camargo, Marlon Dumas, and Oscar Gonz{\'a}lez-Rojas.
\newblock Learning accurate lstm models of business processes.
\newblock In {\em International Conference on Business Process Management},
  pages 286--302. Springer, 2019.

\bibitem[\protect\citeauthoryear{Ceci \bgroup \em et al.\egroup
  }{2014}]{ceci2014completion}
Michelangelo Ceci, Pasqua~Fabiana Lanotte, Fabio Fumarola, Dario~Pietro
  Cavallo, and Donato Malerba.
\newblock Completion time and next activity prediction of processes using
  sequential pattern mining.
\newblock In {\em International Conference on Discovery Science}, pages 49--61.
  Springer, 2014.

\bibitem[\protect\citeauthoryear{Chang \bgroup \em et al.\egroup
  }{2021}]{chang2021sequential}
Jianxin Chang, Chen Gao, Yu~Zheng, Yiqun Hui, Yanan Niu, Yang Song, Depeng Jin,
  and Yong Li.
\newblock Sequential recommendation with graph neural networks.
\newblock In {\em Proceedings of the 44th International ACM SIGIR Conference on
  Research and Development in Information Retrieval}, pages 378--387, 2021.

\bibitem[\protect\citeauthoryear{Evermann \bgroup \em et al.\egroup
  }{2017}]{evermann2017predicting}
Joerg Evermann, Jana-Rebecca Rehse, and Peter Fettke.
\newblock Predicting process behaviour using deep learning.
\newblock {\em Decision Support Systems}, 100:129--140, 2017.

\bibitem[\protect\citeauthoryear{Fang \bgroup \em et al.\egroup
  }{2018}]{fang2018mobile}
Luoyang Fang, Xiang Cheng, Haonan Wang, and Liuqing Yang.
\newblock Mobile demand forecasting via deep graph-sequence spatiotemporal
  modeling in cellular networks.
\newblock {\em IEEE Internet of Things Journal}, 5(4):3091--3101, 2018.

\bibitem[\protect\citeauthoryear{Folino \bgroup \em et al.\egroup
  }{2014}]{folino2014mining}
Francesco Folino, Massimo Guarascio, and Luigi Pontieri.
\newblock Mining predictive process models out of low-level multidimensional
  logs.
\newblock In {\em International conference on advanced information systems
  engineering}, pages 533--547. Springer, 2014.

\bibitem[\protect\citeauthoryear{Garc{\'{\i}}a{-}Dur{\'{a}}n \bgroup \em et
  al.\egroup }{2018}]{garcia2018learning}
Alberto Garc{\'{\i}}a{-}Dur{\'{a}}n, Sebastijan Dumancic, and Mathias Niepert.
\newblock Learning sequence encoders for temporal knowledge graph completion.
\newblock In {\em Proceedings of the 2018 Conference on Empirical Methods in
  Natural Language Processing (EMNLP)}, pages 4816--4821. Association for
  Computational Linguistics, 2018.

\bibitem[\protect\citeauthoryear{Hu \bgroup \em et al.\egroup
  }{2019}]{hu2019graph}
Kun Hu, Zhiyong Wang, Wei Wang, Kaylena A~Ehgoetz Martens, Liang Wang, Tieniu
  Tan, Simon~JG Lewis, and David~Dagan Feng.
\newblock Graph sequence recurrent neural network for vision-based freezing of
  gait detection.
\newblock {\em IEEE Transactions on Image Processing}, 29:1890--1901, 2019.

\bibitem[\protect\citeauthoryear{Jha \bgroup \em et al.\egroup
  }{2019}]{jha2019oulad}
Nikhil~Indrashekhar Jha, Ioana Ghergulescu, and Arghir-Nicolae Moldovan.
\newblock Oulad mooc dropout and result prediction using ensemble, deep
  learning and regression techniques.
\newblock In {\em CSEDU (2)}, pages 154--164, 2019.

\bibitem[\protect\citeauthoryear{Jin \bgroup \em et al.\egroup
  }{2020}]{jin2020recurrent}
Woojeong Jin, Meng Qu, Xisen Jin, and Xiang Ren.
\newblock Recurrent event network: Autoregressive structure inference over
  temporal knowledge graphs.
\newblock In {\em Proceedings of the 2020 Conference on Empirical Methods in
  Natural Language Processing (EMNLP)}, pages 6669--6683, 2020.

\bibitem[\protect\citeauthoryear{Kuzilek \bgroup \em et al.\egroup
  }{2017}]{kuzilek2017open}
Jakub Kuzilek, Martin Hlosta, and Zdenek Zdrahal.
\newblock Open university learning analytics dataset.
\newblock {\em Scientific data}, 4(1):1--8, 2017.

\bibitem[\protect\citeauthoryear{Leontjeva \bgroup \em et al.\egroup
  }{2016}]{leontjeva2016complex}
Anna Leontjeva, Raffaele Conforti, Chiara~Di Francescomarino, Marlon Dumas, and
  Fabrizio~Maria Maggi.
\newblock Complex symbolic sequence encodings for predictive monitoring of
  business processes.
\newblock In {\em International Conference on Business Process Management},
  pages 297--313. Springer, 2016.

\bibitem[\protect\citeauthoryear{Li \bgroup \em et al.\egroup
  }{2021}]{li2021search}
Zixuan Li, Xiaolong Jin, Saiping Guan, Wei Li, Jiafeng Guo, Yuanzhuo Wang, and
  Xueqi Cheng.
\newblock Search from history and reason for future: Two-stage reasoning on
  temporal knowledge graphs.
\newblock In {\em Proceedings of the 59th Annual Meeting of the Association for
  Computational Linguistics and the 11th International Joint Conference on
  Natural Language Processing, {ACL/IJCNLP} 2021, (Volume 1: Long Papers)},
  pages 4732--4743. Association for Computational Linguistics, 2021.

\bibitem[\protect\citeauthoryear{Mauro \bgroup \em et al.\egroup
  }{2019}]{mauro2019activity}
Nicola~Di Mauro, Annalisa Appice, and Teresa Basile.
\newblock Activity prediction of business process instances with inception cnn
  models.
\newblock In {\em International conference of the italian association for
  artificial intelligence}, pages 348--361. Springer, 2019.

\bibitem[\protect\citeauthoryear{Pasquadibisceglie \bgroup \em et al.\egroup
  }{2019}]{pasquadibisceglie2019using}
Vincenzo Pasquadibisceglie, Annalisa Appice, Giovanna Castellano, and Donato
  Malerba.
\newblock Using convolutional neural networks for predictive process analytics.
\newblock In {\em 2019 international conference on process mining (ICPM)},
  pages 129--136. IEEE, 2019.

\bibitem[\protect\citeauthoryear{Reinkemeyer}{2020}]{reinkemeyer2020process}
Lars Reinkemeyer.
\newblock {\em Process Mining in Action}.
\newblock Springer International Publishing, 2020.

\bibitem[\protect\citeauthoryear{Tang \bgroup \em et al.\egroup
  }{2008}]{tang2008arnetminer}
Jie Tang, Jing Zhang, Limin Yao, Juanzi Li, Li~Zhang, and Zhong Su.
\newblock Arnetminer: extraction and mining of academic social networks.
\newblock In {\em Proceedings of the 14th ACM SIGKDD international conference
  on Knowledge discovery and data mining}, pages 990--998, 2008.

\bibitem[\protect\citeauthoryear{Tax \bgroup \em et al.\egroup
  }{2017}]{tax2017predictive}
Niek Tax, Ilya Verenich, Marcello La~Rosa, and Marlon Dumas.
\newblock Predictive business process monitoring with lstm neural networks.
\newblock In {\em International Conference on Advanced Information Systems
  Engineering}, pages 477--492. Springer, 2017.

\bibitem[\protect\citeauthoryear{Taymouri \bgroup \em et al.\egroup
  }{2020}]{taymouri2020predictive}
Farbod Taymouri, Marcello~La Rosa, Sarah Erfani, Zahra~Dasht Bozorgi, and Ilya
  Verenich.
\newblock Predictive business process monitoring via generative adversarial
  nets: the case of next event prediction.
\newblock In {\em International Conference on Business Process Management},
  pages 237--256. Springer, 2020.

\bibitem[\protect\citeauthoryear{Teinemaa \bgroup \em et al.\egroup
  }{2019}]{teinemaa2019outcome}
Irene Teinemaa, Marlon Dumas, Marcello~La Rosa, and Fabrizio~Maria Maggi.
\newblock Outcome-oriented predictive process monitoring: Review and benchmark.
\newblock {\em ACM Transactions on Knowledge Discovery from Data (TKDD)},
  13(2):1--57, 2019.

\bibitem[\protect\citeauthoryear{Tello-Leal \bgroup \em et al.\egroup
  }{2018}]{tello2018predicting}
Edgar Tello-Leal, Jorge Roa, Mariano Rubiolo, and Ulises~M Ramirez-Alcocer.
\newblock Predicting activities in business processes with lstm recurrent
  neural networks.
\newblock In {\em 2018 ITU Kaleidoscope: Machine Learning for a 5G Future (ITU
  K)}, pages 1--7. IEEE, 2018.

\bibitem[\protect\citeauthoryear{Theis and Darabi}{2019}]{theis2019decay}
Julian Theis and Houshang Darabi.
\newblock Decay replay mining to predict next process events.
\newblock {\em IEEE Access}, 7:119787--119803, 2019.

\bibitem[\protect\citeauthoryear{Van Der~Aalst \bgroup \em et al.\egroup
  }{2011a}]{van2011process}
Wil Van Der~Aalst, Arya Adriansyah, Ana Karla~Alves De~Medeiros, Franco
  Arcieri, Thomas Baier, Tobias Blickle, Jagadeesh~Chandra Bose, Peter Van
  Den~Brand, Ronald Brandtjen, Joos Buijs, et~al.
\newblock Process mining manifesto.
\newblock In {\em International Conference on Business Process Management},
  pages 169--194. Springer, 2011.

\bibitem[\protect\citeauthoryear{Van~der Aalst \bgroup \em et al.\egroup
  }{2011b}]{van2011time}
Wil~MP Van~der Aalst, M~Helen Schonenberg, and Minseok Song.
\newblock Time prediction based on process mining.
\newblock {\em Information systems}, 36(2):450--475, 2011.

\bibitem[\protect\citeauthoryear{van Dongen}{2012}]{vandongen_2012}
Boudewijn van Dongen.
\newblock Bpi challenge 2012, Apr 2012.

\bibitem[\protect\citeauthoryear{Vashishth \bgroup \em et al.\egroup
  }{2019}]{vashishth2019composition}
Shikhar Vashishth, Soumya Sanyal, Vikram Nitin, and Partha Talukdar.
\newblock Composition-based multi-relational graph convolutional networks.
\newblock {\em arXiv preprint arXiv:1911.03082}, 2019.

\bibitem[\protect\citeauthoryear{Vaswani \bgroup \em et al.\egroup
  }{2017}]{vaswani2017attention}
Ashish Vaswani, Noam Shazeer, Niki Parmar, Jakob Uszkoreit, Llion Jones,
  Aidan~N Gomez, {\L}ukasz Kaiser, and Illia Polosukhin.
\newblock Attention is all you need.
\newblock In {\em Advances in neural information processing systems}, pages
  5998--6008, 2017.

\bibitem[\protect\citeauthoryear{Wang \bgroup \em et al.\egroup
  }{2020}]{wang2020advanced}
Zhengyang Wang, Meng Liu, Youzhi Luo, Zhao Xu, Yaochen Xie, Limei Wang, Lei
  Cai, Qi~Qi, Zhuoning Yuan, Tianbao Yang, et~al.
\newblock Advanced graph and sequence neural networks for molecular property
  prediction and drug discovery.
\newblock {\em arXiv preprint arXiv:2012.01981}, 2020.

\end{thebibliography}

\end{document}